\documentclass[10pt,twocolumn,letterpaper]{article}

\usepackage[pagenumbers]{wacv} % To force page numbers, e.g. for an arXiv version

\usepackage{graphicx}
\usepackage{amsmath}
\usepackage{amssymb}
\usepackage{booktabs}

\usepackage[ruled,vlined,linesnumbered]{algorithm2e}

\usepackage{xcolor}
 \usepackage{multirow}

\definecolor{commentGreen}{rgb}{0,0.5,0.05}

\usepackage[pagebackref,breaklinks,colorlinks]{hyperref}

\usepackage[capitalize]{cleveref}
\crefname{section}{Sec.}{Secs.}
\Crefname{section}{Section}{Sections}
\Crefname{table}{Table}{Tables}
\crefname{table}{Tab.}{Tabs.}

\def\wacvPaperID{1062} % *** Enter the WACV Paper ID here
\def\confName{WACV}
\def\confYear{2025}

\begin{document}

%%%%%%%%% TITLE - PLEASE UPDATE
\title{Multi-Level Feature Distillation of Joint Teachers\\ Trained on Distinct Image Datasets}

\author{Adrian Iordache, Bogdan Alexe, Radu Tudor Ionescu\\
University of Bucharest, 14 Academiei, 010014, Bucharest, Romania\\
{\tt\small adrian.razvan.iordache@gmail.com, bogdan.alexe@fmi.unibuc.ro, raducu.ionescu@gmail.com}
% For a paper whose authors are all at the same institution,
% omit the following lines up until the closing ``}''.
% Additional authors and addresses can be added with ``\and'',
% just like the second author.
% To save space, use either the email address or home page, not both
% \and
% Second Author\\
% Institution2\\
% First line of institution2 address\\
% {\tt\small secondauthor@i2.org}
}
\maketitle

%%%%%%%%% ABSTRACT
\begin{abstract}
We propose a novel teacher-student framework to distill knowledge from multiple teachers trained on distinct datasets. Each teacher is first trained from scratch on its own dataset. Then, the teachers are combined into a joint architecture, which fuses the features of all teachers at multiple representation levels. The joint teacher architecture is fine-tuned on samples from all datasets, thus gathering useful generic information from all data samples. Finally, we employ a multi-level feature distillation procedure to transfer the knowledge to a student model for each of the considered datasets. We conduct image classification experiments on seven benchmarks, and action recognition experiments on three benchmarks. To illustrate the power of our feature distillation procedure, the student architectures are chosen to be identical to those of the individual teachers. To demonstrate the flexibility of our approach, we combine teachers with distinct architectures. We show that our novel Multi-Level Feature Distillation (MLFD) can significantly surpass equivalent architectures that are either trained on individual datasets, or jointly trained on all datasets at once. Furthermore, we confirm that each step of the proposed training procedure is well motivated by a comprehensive ablation study. We publicly release our code at \url{https://github.com/AdrianIordache/MLFD}.
% \keywords{knowledge distillation \and image classification \and teacher-student model \and multi-level feature distillation}
\end{abstract}

\section{Introduction}
\label{sec:intro}

\begin{figure*}[t]
  \centering
  \includegraphics[width=0.9\textwidth]{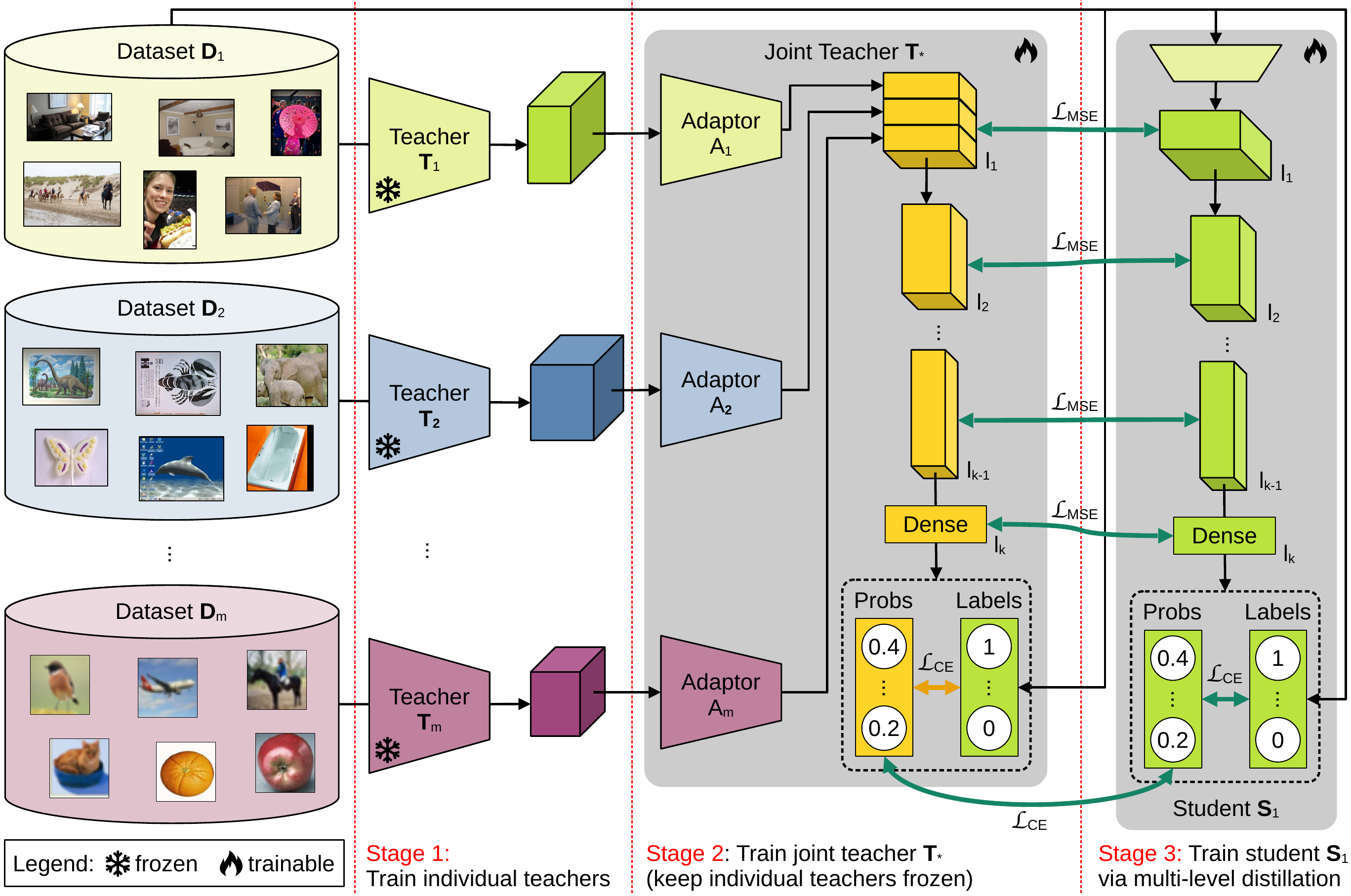}
  \vspace{-0.2cm}
  \caption{Our multi-level feature distillation (MLFD) framework is based on three stages. In the first stage, individual teachers are trained on each dataset. In the second stage, the individual teachers are merged at a certain representation level ($l_1$) into a joint teacher $\textbf{T}_*$, which comprises levels $l_1$, $l_2$, ..., $l_k$. The joint teacher is trained on all datasets $\textbf{D}_1,\textbf{D}_2,...\textbf{D}_m$, while the individual teachers are kept frozen for efficiency reasons. In the third stage, each student $\textbf{S}_i$ is trained via multi-level feature distillation from the joint teacher $\textbf{T}_*$, for all $i \in \{1,2,...,m\}$. To simplify the visualization, only the first student $\textbf{S}_1$ is illustrated in this figure. Best viewed in color.}
  \label{fig_pipeline}
  \vspace{-0.35cm}
\end{figure*}

The usual paradigm in machine learning is to train a model on a single dataset with the desired goal that the trained model should be able to generalize to unseen examples for a specific task. The training algorithm and the dataset, the two key ingredients used in training the model, should be specifically tailored to the task at hand in order to facilitate obtaining a robust model that generalizes to the training data distribution. 
Although the size of the datasets used in training various models for different visual tasks has increased exponentially in the last twenty years, ranging from an order of tens of thousand of images in Caltech-101\cite{feifei-2006-tpami} and PASCAL VOC 2012 \cite{everingham-2010-ijcv} to hundreds of thousands of images in COCO \cite{lin-2014-corr} and millions of images in ImageNet\cite{deng-2009-cvpr}, no dataset is sufficiently rich enough to capture the entire variability and expressiveness of the visual world. Moreover, each dataset comes with an implicit bias introduced by the underlying distribution of the sampled examples it contains \cite{torralba-2011-cvpr}, which might be inherited and even amplified by the trained models \cite{wang-2020-eccv}. Thus, when trained and evaluated on different datasets, models might exhibit a significant difference in performance, showing poor generalization capacity across datasets \cite{torralba-2011-cvpr}. This was even observed for large-scale benchmarks such as ImageNet \cite{Recht-ICML-2019}.
One way to circumvent the problem of overfitting when training on a single dataset and build more robust models is to transfer knowledge from multiple datasets. 
Each dataset comes with its own view of the visual world through its training examples and holds the promise of providing a new and complementary perspective in solving a specific visual task, with respect to other datasets. 
%
%In this work we leverage the knowledge acquired from multiple datasets by first training teacher models from scratch on their own distinct dataset, combining them into a joint teacher, fine-tuned on all datasets,
%
To this end, we propose a Multi-Level Feature Distillation (MLFD) method to distill knowledge from multiple teachers trained on multiple distinct datasets into student models. As illustrated in Figure \ref{fig_pipeline}, our method is composed of three stages. In the first stage, individual teachers are trained from scratch on their own distinct dataset. In the second stage, the individual teachers are combined into a joint teacher, which is trained on all datasets. In the third stage, our method leverages the knowledge learned by the joint teacher by distilling the acquired knowledge from all datasets into dataset-specific student models. The proposed method enables distilling the knowledge using multiple levels of representations, which increases the performance of the student models.

We demonstrate the usefulness of our MLFD method by conducting image classification experiments on seven benchmarks, and action recognition experiments on three benchmarks.
% for the task of image classification on three benchmark datasets: CIFAR-100 \cite{krizhevsky-2009-learning}, ImageNet-Sketch \cite{wang-2019-learning} and TinyImageNet \cite{deng-2009-cvpr} (results on additional datasets are presented in the supplementary). 
A direct way to take advantage of multiple datasets would be to train joint models on all datasets at once. This procedure poses an important challenge, as each dataset defines a different classification task through the distinct class labels associated with each training example. We test two different baselines that mitigate this challenge, showing that our approach significantly surpasses these multi-dataset baselines. Moreover, we thoroughly evaluate our method on both simple and complex architectures.

To summarize, our contributions are the following:
\begin{itemize}
    \item[$\diamond$] \vspace{-0.15cm} We propose a multi-level distillation method that sits at the core of a new teacher-student learning paradigm, where the knowledge acquired by several multiple teachers, grouped into a joint teacher, is distilled into dataset-specific student models. 
    \item[$\diamond$] \vspace{-0.18cm} We show that knowledge from multiple datasets distilled into student architectures brings significant performance gains across multiple architectures, without any extra inference cost.
   % \item[$\diamond$] 
\end{itemize}

%there exist bias in training on a single dataset mitigates this bias by trining on multiple datasets, use knowledge distilation as a vehicle/instrument to mitigate the bias....

\section{Related Work}
\label{sec:related-work}

\noindent
{\bf Knowledge distillation.} The goal of Knowledge Distillation (KD) is to transfer knowledge from a large teacher model to a small student model, such that the obtained student model mimics the behavior of the larger model \cite{hinton-2014-nips,lopez2015unifying,mirzadeh2020improved,Cho_2019_ICCV,ji2020knowledge, cheng2020explaining}. The work of~\cite{hinton-2014-nips} defines the knowledge in the form of the output of the large model (logits). The basic form of knowledge distillation consists in training the small student model to reproduce the logits of the large teacher model. This can be done either in the unsupervised \cite{bai2023masked,lao2023masked} or supervised \cite{Lin-CVRP-2023} scenario. In the supervised scenario, with ground-truth labels available at training time, Hinton \etal~\cite{hinton-2014-nips} show that significant improvements can be obtained by minimizing an objective function that takes into account the cross-entropy between the logits of the two models, but also another term that enforces the student model to predict ground-truth labels. Other works \cite{romero-2015-iclr,chen-2021-cvpr,xu-2020-eccv} consider knowledge at the feature level, employing distillation by matching feature distributions as well as logit outputs between the teacher and student models. Romero \etal~\cite{romero-2015-iclr} use intermediate-level hints from the teacher hidden layers to guide the training process of the student, enforcing a deeper and thinner student network to learn an intermediate representation that is predictive of the intermediate representations of the wide and shallower (but still deep) teacher network. 
Park \etal~\cite{park-2019-cvpr} transfer knowledge in the form of mutual relations of data examples, by including distance-wise and angle-wise distillation losses that penalize structural differences in relations. 
%bogdan - added this two citations
%https://link.springer.com/chapter/10.1007/978-3-031-19809-0_20
Li~\cite{Li-ECCV-2022} reuses channel-wise and layer-wise meaningful features within the student to provide teacher-like knowledge without an additional model, in a teacher-free feature distillation framework. Liu \etal\cite{LIU-ICLR-2023} propose a two-stage knowledge distillation method, which relies on a simple Feature Transform module consisting of two linear layers.
Our proposed multi-level feature distillation method considers knowledge both at the logits level and feature level, but, unlike other approaches, it uses multiple teachers, each of them trained on a distinct dataset. 
Usually, knowledge distillation is applied for model compression~\cite{hinton-2014-nips}, where the teacher model has a much larger capacity and memory footprint w.r.t.~the student model. Our method allows student architectures to be identical to those of the individual teachers. Moreover, our method is very flexible, since it can combine teachers with distinct architectures. 

%bogdan - added this three citations
A different strand of work uses search techniques, such as Monte Carlo tree search~\cite{Li-ICCV-2023} or evolutionary search~\cite{Li-NIPS-2023,Dong-CVPR-2023}, to investigate optimal knowledge distillation designs for distilling a given teacher-student. These approaches are orthogonal to our work, and their combination can further improve the result of knowledge distillation.

\noindent
{\bf Dataset distillation.} The goal of dataset distillation\cite{wang-2018-arxiv,cazenavette_2022_cvpr} is to synthesize an original large dataset with a small dataset, such that models trained on the smaller dataset exhibit similar performance to models trained on the original one. Although our framework enables using multiple distinct datasets, we do not aim to compress these datasets. Instead, we distill knowledge from a joint teacher obtained by combining several individual teachers, each trained on a distinct dataset. This is an orthogonal approach to dataset distillation.

%This is something orthogonal to knowledge distillation. 
%https://arxiv.org/pdf/2301.07014.pdf
%https://openaccess.thecvf.com/content/CVPR2023/papers/Jin_Multi-Level_Logit_Distillation_CVPR_2023_paper.pdf

%{\bf Relation to teacher-student learning.} => this still KD
%{\bf Multi-layer Knowledge Distillation} 
% see this https://www.sciencedirect.com/science/article/abs/pii/S0952197624003282
%-------------------------------------------------------------------------
\section{Method}
\label{sec:method}

We propose a new multi-level distillation method to improve the generalization capacity and performance of multiple models trained on distinct datasets, at no additional costs during inference time. We achieve this by combining multiple teacher models, each trained on a distinct dataset, into a joint teacher, which is trained on data samples from the combined datasets. %The teachers are fused at multiple depth levels, allowing the joint teacher to learn multiple levels of joint representations. 
Then, the knowledge from the joint teacher is distilled into dataset-specific students. The distillation step is carried out at multiple representation levels, resulting in a multi-level feature distillation (MLFD) framework. Not only do the students receive information from different teachers, but they also capture information from distinct datasets, which significantly boosts their generalization capacities. Our MLFD framework is formally presented in Algorithm \ref{alg_MLFD}.

\begin{algorithm*}[!t]
\small{
\caption{Multi-Level Feature Distillation}
\label{alg_MLFD} 
\KwIn{$\mathcal{D}=\{ \textbf{D}_i \vert \textbf{D}_i=\{ (\boldsymbol{I}^i_1, y^i_1), (\boldsymbol{I}^i_2,y^i_2) .., (\boldsymbol{I}^i_{n_i},y^i_{n_i}) \}, \forall i \in \{1,2,...,m\}\}$ - the set of training sets of labeled images, $\mathcal{T}=\{\textbf{T}_1, \textbf{T}_2,...,\textbf{T}_m\}$ - the set of teacher models, $\mathcal{S}=\{\textbf{S}_1, \textbf{S}_2,...,\textbf{S}_m\}$ - the set of student models, $\eta_{\textbf{T}_i}$ - the learning rate of teacher $\textbf{T}_i$, $\eta_{\textbf{T}_*}$ - the learning rate of the joint teacher, $\eta_{\textbf{S}_i}$ - the learning rate of the student $\textbf{S}_i$, $\textbf{L}=\{l_1,l_2,...,l_k\}$ - the set of layer indexes at which the multi-level fusion and distillation is performed.}

\KwOut{$\theta_{\textbf{S}_i}, \forall i \in \{1,2,...,m\}$ - the trained weights of each student model $\textbf{S}_i$.}

\ForEach{$i \in \{1,2,...,m\}$}{
$\theta_{\textbf{T}_i} \sim \mathcal{N}\left(0,\frac{2}{d_{in_i} + d_{out_i}}\right);$ \textcolor{commentGreen}{$\lhd$ initialize weights of teacher $\textbf{T}_i$ using Xavier \cite{Glorot-AISTATS-2010}}\\

\Repeat{convergence}{
        \ForEach{$j \in \{1,2,...,n_i\}$}{
            $t_j \leftarrow \textbf{T}_i(\boldsymbol{I}^i_j, \theta_{\textbf{T}_i});$ \textcolor{commentGreen}{$\lhd$ get class probabilities predicted by teacher $\textbf{T}_i$}\\
            $\theta_{\textbf{T}_i} \leftarrow \theta_{\textbf{T}_i} - \eta_{\textbf{T}_i} \cdot \nabla \mathcal{L}_{\scriptsize{\mbox{CE}}}(y^i_j, t_j);$ \textcolor{commentGreen}{$\lhd$ train the teacher $\textbf{T}_i$}\\
        }
    }
}
    
$\textbf{T}_* \gets \mbox{\emph{fuse}}\left(\mathcal{T}, \textbf{L} \right);$ \textcolor{commentGreen}{$\lhd$ apply the fusion in Eq.~\eqref{eq_fuse_multi} to obtain joint teacher}\\  

\Repeat{convergence}{
    \ForEach{$i \in \{1,2,...,m\}$}{
        \ForEach{$j \in \{1,2,...,n_i\}$}{
            $t_j \gets \textbf{T}_*(\boldsymbol{I}^i_j, \theta_{\textbf{T}_*});$ \textcolor{commentGreen}{$\lhd$ get class probabilities predicted by teacher $\textbf{T}_*$}\\
            $\theta_{\textbf{T}_*} \gets \theta_{\textbf{T}_*} - \eta_{\textbf{T}_*} \cdot \nabla \mathcal{L}_{\scriptsize{\mbox{CE}}}(y^i_j, t_j);$ \textcolor{commentGreen}{$\lhd$ train the joint teacher $\textbf{T}_*$}\\
        }
    }
}

\ForEach{$i \in \{1,2,...,m\}$}{
$\theta_{\textbf{S}_i} \sim \mathcal{N}\left(0,\frac{2}{d_{in_i} + d_{out_i}}\right);$ \textcolor{commentGreen}{$\lhd$ initialize weights of student $\textbf{S}_i$ using Xavier \cite{Glorot-AISTATS-2010}}\\

\ForEach{$j \in \{1,2,...,n_i\}$}{
            $t_j, \boldsymbol{e}^{\textbf{T}^{l_1}_*}_j,...,\boldsymbol{e}^{\textbf{T}^{l_k}_*}_j \gets \textbf{T}_*(\boldsymbol{I}^i_j, \theta_{\textbf{T}_*});$ \textcolor{commentGreen}{$\lhd$ get class probabilities and multi-level embeddings from joint teacher}\\
}
\Repeat{convergence}{
        \ForEach{$j \in \{1,2,...,n_i\}$}{
            $s_j, \boldsymbol{e}^{\textbf{S}^{l_1}_i}_j,...,\boldsymbol{e}^{\textbf{S}^{l_k}_i}_j \gets \textbf{S}(\boldsymbol{I}^i_j, \theta_{\textbf{S}_i});$ \textcolor{commentGreen}{$\lhd$ get class probabilities and multi-level embeddings from student $\textbf{S}_i$}\\
            $\mathcal{L}_{\scriptsize{\mbox{KD}}} \gets \mathcal{L}_{\scriptsize{\mbox{CE}}}(y^i_j, s_j) + \alpha \cdot \mathcal{L}^{\textbf{T}_*}_{\scriptsize{\mbox{CE}}}(t_j, s_j) + \sum_{c=1}^k \beta_{c} \cdot \mathcal{L}^{\textbf{T}_*}_{\scriptsize{\mbox{MSE}}}\left(\boldsymbol{e}^{\textbf{T}^{l_c}_*}_j, \boldsymbol{e}^{\textbf{S}^{l_c}_i}_j\right)\!;$ \textcolor{commentGreen}{$\lhd$ apply Eq.~\eqref{eq_loss_KD}}\\
            $\theta_{\textbf{S}_i} \gets \theta_{\textbf{S}_i} - \eta_{\textbf{S}_i} \cdot \nabla \mathcal{L}_{\scriptsize{\mbox{KD}}};$ \textcolor{commentGreen}{$\lhd$ train the student using the joint loss}\\
        }
    }
}
}
\end{algorithm*}

The MLFD algorithm essentially takes as input a set of datasets $\mathcal{D}$, a set of teacher architectures $\mathcal{T}$, and a set of student architectures $\mathcal{S}$, where $|\mathcal{D}|=|\mathcal{T}|=|\mathcal{S}|=m$. We underline that each dataset can comprise a different number of images. Moreover, the teacher and student architectures can be diverse, which enables the use of a suitable architecture for each specific dataset. There are some additional hyperparameters given as input, such as the learning rates and the set of layer indexes $\textbf{L}$ at which the multi-level fusion and distillation is performed.

In steps 1-7, we train the set of teachers from scratch via a variant of stochastic gradient descent (SGD), such that each teacher is trained on its own dataset. Since our algorithm does not impose the use of the same architecture for every teacher, we are free to harness existing pre-trained architectures from the public web domain. This means that the individual teacher training step can be omitted, whenever this is desired.

In step 8, the individually trained teachers are fused into a joint architecture, denoted as $\textbf{T}_*$. The fusion is performed at a certain representation level, denoted as $l_1$. To simplify the notations, we use $l_1$ to denote the index of the layer that is fused from each teacher, disregarding the fact that $l_1$ can actually represent a different index for each teacher architecture. Formally, the \emph{fuse} function used in step 8 of Algorithm \ref{alg_MLFD} is defined as follows:
\begin{equation}\label{eq_fuse_multi}
\textbf{T}_*\!=\!\mbox{NN}^{l_k}\!\!\left(...\;\mbox{NN}^{l_2}\!\!\left(\mbox{NN}^{l_1}\!\!\left(\boldsymbol{e}^{\textbf{T}^{l_1}_*}_j\!,\theta^{l_1}_{\textbf{T}_*}\right)\!,\theta^{l_2}_{\textbf{T}_*}\right)...,\theta^{l_k}_{\textbf{T}_*}\right)\!,
\end{equation}
where $\mbox{NN}^{l_c}$ is a neural network block that transforms the features at layer $l_c$ into a set of features that are given as input to layer $l_{c+1}$, $\theta^{l_c}_{\textbf{T}_*}$ are the features of the neural block $\mbox{NN}^{l_c}$, and $j$ iterates through all images in $\mathcal{D}$. We note that the neural blocks $\mbox{NN}^{l_c}$ are typically shallow, comprising either a conv and a pooling operation, or one or two dense layers. The type of block (convolutional or dense) depends on the desired architectural design of the joint teacher. The embedding $\boldsymbol{e}^{\textbf{T}^{l_1}_*}_j$ given as input to layer $l_1$ of the joint teacher $\textbf{T}_*$, corresponding to the $j$-th training image, is computed as follows:
\begin{equation}\label{eq_fuse_one}
\resizebox{0.902\columnwidth}{!}{%
$ \!\!\!\boldsymbol{e}^{\textbf{T}^{l_1}_*}_j\!\!=\!\left[A_1\!\!\left(\boldsymbol{e}^{\textbf{T}^{l_1}_1}_j\!,\theta_{A_1}\right)\!,A_2\!\!\left(\boldsymbol{e}^{\textbf{T}^{l_1}_2}_j\!,\theta_{A_2}\!\right)\!,...,A_m\!\!\left(\boldsymbol{e}^{\textbf{T}^{l_1}_m}_j,\theta_{A_m}\right)\!\right]\!,$
}
\end{equation}
where $[\cdot,\cdot]$ is the channel-wise concatenation operation, $\boldsymbol{e}^{\textbf{T}^{l_1}_1}_j, \boldsymbol{e}^{\textbf{T}^{l_1}_2}_j, ...,\boldsymbol{e}^{\textbf{T}^{l_1}_m}_j$ are the embeddings returned by the individual teachers for the $j$-th image, and $A_1, A_2, ..., A_m$ are adaptor blocks that bring the features at layer $l_1$ from all teachers to the same spatial dimensions, so that they can be concatenated. Each adaptor block $A_i$ is parameterized by the weights $\theta_{A_i}$, $\forall i \in \{1,2,...,m\}$.

In steps 9-14, we train the joint teacher on all datasets, via SGD. We note that the trainable weights $\theta_{\textbf{T}_*}$ of the joint teacher are composed of the weights that correspond to the neural blocks $\mbox{NN}^{l_c}$, denoted by $\theta^{l_c}_{\textbf{T}_*}$, and the weights that correspond to the adaptor blocks $A_i$, denoted by $\theta_{A_i}$. In other words, the weights of individual teachers that precede the fusion layer are kept frozen. This allows us to extract the features $\boldsymbol{e}^{\textbf{T}^{l_1}_i}_j$ of the $j$-th training image in one go, before training the joint teacher, which significantly reduces the training time and the memory footprint, while training the joint teacher.
% In general, $l_c$ denotes the index of the $c$-th layer that is fused from each teacher, where $c \in \{1,2,...,k\}$.

After training the joint teacher, it is time to distill its knowledge into the students. Our multi-level distillation is performed in steps 15-24 of Algorithm \ref{alg_MLFD}. In step 16, we initialize the weights of the student $\textbf{S}_i$. Then, in steps 17-18, we extract the embeddings at multiple representation levels from the joint teacher $\textbf{T}_*$, where the chosen representation levels are specified through the set $\textbf{L}$. The embeddings $\boldsymbol{e}^{\textbf{T}^{l_c}_*}_j$ returned by the joint teacher in step 18 represent the features used inside the block $\mbox{NN}^{l_c}$, where $c \in \{1,2,...,k\}$. Finally, in steps 19-24, we distill the knowledge learned by the joint teacher into the student $\textbf{S}_i$. The student is trained end-to-end (there are no frozen weights) via SGD. To perform the multi-level knowledge distillation, we employ the following loss function:
\begin{equation}\label{eq_loss_KD}
\resizebox{0.905\columnwidth}{!}{%
$\!\!\!\!\mathcal{L}_{\scriptsize{\mbox{KD}}} = \mathcal{L}_{\scriptsize{\mbox{CE}}}(y^i_j, s_j) + \alpha\!\cdot\!\mathcal{L}^{\textbf{T}_*}_{\scriptsize{\mbox{CE}}}(t_j, s_j) + \sum_{c=1}^k \beta_{c}\!\cdot\!\mathcal{L}^{\textbf{T}_*}_{\scriptsize{\mbox{MSE}}}\left(\boldsymbol{e}^{\textbf{T}^{l_c}_*}_j\!, \boldsymbol{e}^{\textbf{S}^{l_c}_i}_j\!\right)\!,$
 }
\end{equation}
where $\mathcal{L}_{\scriptsize{\mbox{CE}}}$ is the cross-entropy (CE) loss between the ground-truth one-hot class label $y^i_j$ and the class probabilities $s_j$ predicted by the student, $\mathcal{L}^{\textbf{T}_*}_{\scriptsize{\mbox{CE}}}$ is the cross-entropy loss between the class probabilities $t_j$ returned by the joint teacher and those of the student, and $\mathcal{L}^{\textbf{T}_*}_{\scriptsize{\mbox{MSE}}}$ is the mean squared error (MSE) between the embedding $\boldsymbol{e}^{\textbf{T}^{l_c}_*}_j$ at layer $l_c$ returned by the joint teacher and the embedding $\boldsymbol{e}^{\textbf{S}^{l_c}_i}_j$ at layer $l_c$ returned by the student, respectively. In general, $\boldsymbol{e}^{\textbf{T}^{l_c}_*}_j$ might not have compatible dimensions with $\boldsymbol{e}^{\textbf{S}^{l_c}_*}_j$. When this happens, an adaptor layer can be inserted in order to bring the shape of $\boldsymbol{e}^{\textbf{T}^{l_c}_*}_j$ to the desired size. The hyperparameters $\alpha \geq 0$ and $\beta_c \geq 0$ control the importance of the distillation losses with respect to the classification loss. Note that the loss at a certain representation level $c$ has its own contribution to the overall loss defined in Eq.~\eqref{eq_loss_KD}, which is determined by $\beta_c$. For optimal results, these hyperparameters can be fine-tuned on the validation set of each dataset $\textbf{D}_i \in \mathcal{D}$. The algorithm returns the weights $\theta_{\textbf{S}_i}$ for each student $\textbf{S}_i \in \mathcal{S}$.

\section{Experiments}
\label{sec:experiments}

% We evaluate our method for the task of image classification.  

\subsection{Datasets}

We next present experiments on three datasets for image classification: CIFAR-100 \cite{krizhevsky-2009-learning}, ImageNet-Sketch \cite{wang-2019-learning} and TinyImageNet \cite{deng-2009-cvpr}. Results on additional datasets and tasks are discussed in the supplementary.
%We employ our method on three datasets for image classification: CIFAR-100 \cite{krizhevsky-2009-learning}, ImageNet-Sketch \cite{wang-2019-learning} and TinyImageNet \cite{deng-2009-cvpr}.
%In our case, the datasets have a fraction of joint classes, but from our intuition, this does not represent the requirement for our method. 

%CIFAR-100 \cite{krizhevsky-2009-learning}, ImageNet-Sketch \cite{wang-2019-learning} and TinyImageNet \cite{deng-2009-cvpr}

\noindent
\textbf{CIFAR-100.} 
The CIFAR-100 dataset consists of $60,000$ color images of $32 \times 32$ pixels, grouped into $100$ classes. % Furthermore, these $100$ classes are grouped in $20$ superclasses, for example the classes {\emph {crocodile}}, {\emph{dinosaur}}, {\emph{lizard}}, {\emph{turtle}}, and {\emph{snake}} belong to the superclass {\emph{reptile}}.  
We use the official split with $500$ training images and $100$ testing images per class, for a total of $50,000$ training images and $10,000$ test images.

\noindent
\textbf{ImageNet-Sketch.} The ImageNet-Sketch dataset consists of black and white images from $1000$ classes, corresponding to the ImageNet validation classes. The dataset is intended to be used for evaluating the ability of models to generalize to out-of-domain data, represented by sketch-like images. The dataset comprises $41,088$ training images and $10,752$ test images. 

\noindent
\textbf{TinyImageNet.} The TinyImageNet dataset is a subset of the ImageNet~\cite{deng-2009-cvpr} dataset, containing $100,000$ training images from $200$ object classes. All images have a resolution of $64 \times 64$ pixels. % and $10,000$ validations and $10,000$ test images. 
As the labels for official test images are not publicly available, we use the $10,000$ validations images as test data. 
%https://paperswithcode.com/dataset/tiny-imagenet

%The efficiency of our method is due to the different perspectives of the shared features across datasets, providing better generalization, and not necessarily restricted to only the joint classes. A factor we consider to be impactful to our method is the ratio of training samples from each dataset. 

%In our case, we selected the datasets considering that CIFAR-100 and ImageNet-Sketch have roughly the same number of training samples, 50 thousand samples, and TinyImageNet has approximately double the number of samples. 

%The final reason for selecting these datasets was to see the impact of features shared from datasets containing colors such as CIFAR-100 and TinyImageNet to ImageNet-Sketch, a dataset with black and white images.

\subsection{Baselines}
%We test our knowledge distillation method, we proposed three baselines:
In our experiments, we consider three types of baselines that are described below.
% \begin{itemize}
%     \item \textbf{Dataset-specific models:} As the primary baseline for our method, we consider each initial trained model, the individual teacher, which does not share information across datasets and will be used to generate the joint teacher.
%     \item \textbf{Multi-head models with sequential batch training:} As a first baseline for joint dataset training, we proposed an architecture composed of a shared backbone and an independent prediction head for each dataset. During every iteration, we train our model by forwarding sequentially a batch from every dataset, weighting equally each batch in the final loss, and accumulating the gradients from all batches to be updated at once.
%     \item \textbf{Joint-head models with mixed batch training:} A second approach for joint dataset training, it is represented by a model with a single head for predictions, containing the total number of classes across all datasets. During training, each batch will contain a mixture of images from all the available datasets, with the label shifted accordingly.
% \end{itemize}

\noindent\textbf{Dataset-specific models.} As primary baselines for our method, we consider models that have identical architectures to our student models. These baseline models are trained from scratch on individual datasets, so they do not benefit from information gathered from other datasets. An accuracy boost over these baselines will demonstrate the benefit of our multi-dataset distillation approach.

\noindent\textbf{Multi-head multi-dataset models.} As a second baseline, we propose an architecture composed of a shared backbone and an independent prediction head for each dataset. During every iteration, we train our model by sequentially forwarding a batch from every dataset, and accumulating the gradients from all batches to be updated at once. Each batch is equally weighted in the final loss.

\noindent\textbf{Joint-head multi-dataset models.} The third baseline is an alternative  approach for joint dataset training. It is represented by a model with a single classification head, containing the total number of classes across all datasets. During training, each batch contains a mixture of images from all the available datasets, where each label is a one-hot vector over the joint set of classes from all datasets.

\subsection{Experimental Setup} 

We evaluate our method on two sets of individual teachers, denoted as $\mathcal{T}_1$ and $\mathcal{T}_2$, respectively. $\mathcal{T}_1$ contains three models, namely a ResNet-18 \cite{he-2016-cvpr} trained on CIFAR-100, an EfficientNet-B0 \cite{tan-2019-plmr} trained on TinyImageNet, and a SEResNeXt-26D \cite{hu-2018-cvpr} trained on ImageNet-Sketch. $\mathcal{T}_2$ contains a different lineup of models, namely a ConvNeXt-V2 Tiny \cite{woo_2023_cvpr} trained on CIFAR-100, a SwinTransformer-V2 Tiny \cite{liu_2022_cvpr} trained on TinyImageNet, and a FastViT SA24 \cite{vasu_2023_iccv} trained on ImageNet-Sketch. In summary, $\mathcal{T}_1$ and $\mathcal{T}_2$ contain a variety of state-of-the-art image classification models, being based on both convolutional and transformer architectures. This is to demonstrate that the individual teachers can have distinct architectures.
To demonstrate the full power of our multi-dataset distillation pipeline, we train the individual teachers from scratch and refrain from using weights pre-trained on ImageNet. Moreover, the student models, as well as the dataset-specific baseline models, have identical architectures to the individual teachers in $\mathcal{T}_1$ and $\mathcal{T}_2$, respectively. This allows us to attribute the reported accuracy improvements to our method, and not to the architectural choices.

We run the experiments on a machine with an NVIDIA GeForce RTX 3090 GPU, an Intel i9-10940X 3.3 GHz CPU, and 128 GB of RAM. Due to hardware limitations, we %prefer using variations of lightweight models, as well as 
employ techniques such as gradient accumulation, mixed precision, and offline distillation. For the same reason, we cache the embeddings from the layers at index $l_1$ of the individual teachers, before training the joint teacher model.

%The \textbf{\textit{proof-of-concept}} set composed by ResNet-18 trained on CIFAR-100, EfficientNet-B0  trained on TinyImageNet, and SEResNeXt-26D on ImageNet-Sketch. 

%The \textbf{\textit{state-of-the-art}} set composed by ConvNeXt-V2 Tiny trained on CIFAR-100, SwinTransformer-V2 Tiny trained on TinyImageNet, and FastViT SA24 on ImageNet-Sketch.

% \begin{enumerate}
%     \item The \textbf{\textit{proof-of-concept}} set composed by: ResNet-18 for CIFAR-100, EfficientNet-B0 for TinyImageNet, SEResNeXt-26D for ImageNet-Sketch.
%     % \begin{itemize}
%     %     \item ResNet-18 for CIFAR-100
%     %     \item EfficientNet-B0 for TinyImageNet
%     %     \item SEResNeXt-26D for ImageNet-Sketch
%     % \end{itemize}
%     % \item The \textbf{\textit{state-of-the-art}} set: 
%     % \begin{itemize}
%     %     \item ConvNeXt-V2 Tiny for CIFAR-100
%     %     \item SwinTransformer-V2 Tiny for TinyImageNet
%     %     \item FastViT SA24 for ImageNet-Sketch
%     % \end{itemize}
    
% \end{enumerate}

\begin{table*}[!h]
% \vspace{-0.2cm}
% \resizebox{\columnwidth}{!}{%
\centering
% \scriptsize{
\begin{tabular}{|l|c|c|c|c|c|c|c|}
\hline
\multirow{2}{*}{Architecture}            & Learning   & Batch & Stochastic & \multirow{2}{*}{Dropout} & Weight & \multirow{2}{*}{Temperature} & $\alpha$ and $\beta_c$ \\ 
            & rate  & size & depth &  & decay &  &  coefficients \\ 

\hline
\hline
ResNet-18               & 1e-4 & 768        & 0                & -       & 0.001        & 2           & 0.6, 0.2, 0.2      \\ \hline
EfficientNet-B0         & 1e-4 & 240        & 0.2              & 0.2     & 0.001        & 2           & 0.6, 0.2, 0.2      \\ \hline
SEResNeXt-26D           & 1e-4 & 192        & 0                & -       & 0.001        & 2           & 0.6, 0.2, 0.2      \\ \hline
ConvNeXt-V2 Tiny        & 1e-5 & 96         & 0.6              & -       & 0.5          & 5           & 0.6, 02, 0.2       \\ \hline
SwinTransformer-V2 Tiny & 4e-4 & 176        & 0.6              & -       & 0.5          & 2           & 0.6, 0.2, 0.2      \\ \hline
FastViT SA24            & 1e-4 & 102        & 0.3              & 0.2     & 0.5          & 2           & 0.4, 0.4, 0.2      \\ \hline
\end{tabular}
% }
\vspace{-0.2cm}
\caption{Optimal hyperparameter settings for the various neural architectures used in our experiments.}\label{tab_params}
% \vspace{-0.2cm}
\end{table*}

\subsection{Hyperparameter Tuning}

We train each model from $\mathcal{T}_1$ and $\mathcal{T}_2$ in a different manner, adapted to the model and the target dataset. For models in $\mathcal{T}_1$, we apply a training procedure based on a constant learning rate, using the Adam optimizer \cite{Kingma-ICLR-1015} with weight decay for regularization, and the AutoAugment~\cite{cubuk_2019_cvpr} with ImageNet policy for augmentation. We tried replacing the constant learning rate with one-cycle learning for all models in $\mathcal{T}_1$, without observing any significant improvements. All models are trained for $200$ epochs on images with a resolution of $224 \times 224$ pixels.

%For the training settings of both sets of models, we apply different strategies. In the proof-of-concept set, we apply a simplistic training setting, replacing one-cycle learning with a constant learning rate, as no significant improvement resulted by one-cycle, using a regular Adam optimizer with weight decay for regularization, and the AutoAugment with ImageNet policy for augmentation. All models are being trained for 200 epochs on 224 image resolution.

\begin{table*}[t!]
% \vspace{-0.2cm}
\centering
% \resizebox{\columnwidth}{!}{%
\begin{tabular}{|c|c|cc|cc|cc|}
    \hline
     \multirow{2}{*}{{Model}} & {Distillation} & \multicolumn{2}{c|}{{CIFAR-100}} & \multicolumn{2}{c|}{{TinyImageNet}} & 
     \multicolumn{2}{c|}{{ImageNet-Sketch}} \\
     
     & {levels} & \texttt{acc@1} & \texttt{acc@5} 
     & \texttt{acc@1} & \texttt{acc@5} 
     & \texttt{acc@1} & \texttt{acc@5} \\
     
     \hline
     \hline
     
     %\textbf{Dataset-specific models}
     {Dataset-specific models} & -
     
     & $55.07$ & $81.62$ 
     & $45.09$ & $70.52$ 
     & $49.67$ & $71.39$ \\ 

     %\textbf{Multi-head models} 
     {Multi-head model} & -
     
     & $36.61$ & $65.00$ 
     & $35.26$ & $63.64$ 
     & $21.19$ & $43.47$ \\

      %\textbf{Joint-head models} 
      Joint-head model & -
     & $59.46$ & $85.98$ 
     & $45.13$ & $73.09$ 
     & $41.59$ & $66.79$ \\
      \hline
     
      %\textbf{$L_1$ Students} 
     \multirow{2}{*}{Students (our)} & $\textbf{L}_1$
     & $58.65$ & $83.97$  
     & $50.86$ & $76.07$  
     & $51.64$ & $72.81$  \\ 

     %\textbf{$L_2$ Students} 
     & $\textbf{L}_2$ 
     & $\mathbf{62.25}$ & $\mathbf{84.64}$  
     & $\mathbf{51.61}$ & $\mathbf{76.46}$  
     & $\mathbf{61.31}$ & $\mathbf{78.36}$  \\ 
     
     \hline
     
      %\textbf{$L_1$ Joint Teacher} 
     \multirow{2}{*}{Joint teacher (ours)} & $\textbf{L}_1$
     & $61.00$ & $85.44$  %cifar100
     & $46.97$ & $\mathbf{72.53}$   %tiny-imagenet
     & $56.35$ & $75.60$  \\ %ucf 

      %\textbf{$L_2$ Joint Teacher}
    & $\textbf{L}_2$ 
     & $\mathbf{63.34}$ & $\mathbf{87.03}$  %cifar100
     & $\mathbf{47.78}$ & $71.71$  %tiny-imagenet
     & $\mathbf{61.45}$ & $\mathbf{79.38}$  \\ %ucf 
     
     \hline
     
\end{tabular}
% }
\vspace{-0.2cm}
\caption{Results for the set $\mathcal{T}_1$, including the three baselines, the students obtained by employing multi-level distillation using embeddings extracted at one ($\textbf{L}_1$) or two levels ($\textbf{L}_2$), and the corresponding joint teachers. The results of the best student and the best teacher are highlighted in bold.}
\label{tab:T1}
\vspace{-0.2cm}
\end{table*}

For models in $\mathcal{T}_2$, each architecture has a specific training recipe described in the corresponding original papers. We start with almost the same hyperparameters, and make minimal changes in order to adapt the hyperparameters to each dataset, aiming for a better performance. Several choices led to significant improvements of each model: adding stochastic depth, label smoothing, and dropout in the final layers. The models in $\mathcal{T}_2$ are trained with AdamW \cite{Loshchilov-ICLR-2019} and weight decay. We tried several well-established ImageNet augmentation techniques such as RandAugment \cite{cubuk-2020-nips}, MixUp \cite{zhang-2018-iclr}, CutMix \cite{yun_2019_iccv}, and RandomErasing \cite{zhong-2020-aaai}, but all of them led to performance drops. Thus, we settle for using the AutoAugment technique. To maintain compatibility with the included transformer models, we use $256 \times 256$ input images for all models in $\mathcal{T}_2$. Regarding the learning rate schedule, the one-cycle procedure %near zero final learning 
provides a boost in performance for the second set of models, but downgrades the quality of fused features leading to worse performance in the joint teacher. For that reason, we either maintain the learning rate constant or employ a one-cycle learning rate procedure with the final learning rate having the same value as the constant one.
%For the final setting, we add stochastic depth leading to significant improvements for each model, we also add label smoothing and dropout in the final layer, using a higher weight decay than in previous set models, we move from Adam to AdamW. We test established ImageNet augmentation techniques such as RandAugment, MixUp, CutMix, and RandomErasing, leading to a downgrade in performance, we maintain the AutoAugment. We switch to 256 image resolution to maintain compatibility with the transformer models. As for the learning rate schedule, the one-cycle procedure near zero final learning provides a boost in performance in this set of models but downgrades the quality of fused features leading to worse performance in the joint teacher. For that reason, we either maintain the learning rate constant or employ a one-cycle learning rate procedure with the final learning rate having the same value as the constant one.

In Table \ref{tab_params}, we report specific values of optimal hyperparameter choices. All other hyperparameters are set to their default values.

% \begin{table*}[!t]
% \centering
% \resizebox{\columnwidth}{!}{%
%   \begin{tabular}{|c|c|c|c|c|c|c|c|}
%     Architecture  &  LR   &  Batch Size  & Stochastic Depth & Dropout & Weight Decay  & Temp & Distillation Coefs \\   
%     Resnet-18 

%   \end{tabular}%
% }
%       % \vspace{-0.2cm}
%       \caption{Optimal hyperparameter settings for the various neural architectures used in our experiments.}\label{tab_vision}
% \end{table*}

\subsection{Joint Teacher Architecture}

We explore several architectures for the joint teacher, depending on the type of layers that need to be fused.
When combining embeddings from dense layers of individual teachers, we initially explored variations of feed-forward layers, 1D convolutional layers, and multi-head attention layers. For each alternative, we considered versions with and without residual connections. However, the best results were obtained by using a single highly regularized linear layer to project the concatenated embedding vectors from our individual teachers. To regulate the single layer, we employ dropout with a probability ranging between 0.8 and 0.9. 
Mixing embeddings from convolutional or transformer layers of individual teachers is typically performed via pointwise convolutional layers. Subsequent layers in the joint teacher follow a basic architecture comprising convolutional layers, batch normalization, and adaptive average pooling. Gaussian error linear unit (GeLU) is the activation function employed in all neurons. A dense layer with a high dropout rate is added just before the classification head.

In general, the depth of the trainable joint teacher is chosen such that, when combined with the frozen layers of the individual teachers, it results in a depth that roughly matches that of the individual teachers. Although a deeper joint teacher might bring additional performance gains, we refrain from using deeper or complex joint teachers to ensure that the reported performance gains are due to the proposed training procedure, and not due to the higher capability of the joint teacher architecture. 

% For downsampling, regular convolution layers and pooling are applied.

\begin{table*}[t!]
% \vspace{-0.2cm}
\centering
% \resizebox{\columnwidth}{!}{%
\begin{tabular}{|c|c|cc|cc|cc|}
    \hline
      \multirow{2}{*}{{Model}} & {Distillation} & \multicolumn{2}{c|}{{CIFAR-100}} & \multicolumn{2}{c|}{{TinyImageNet}} & 
     \multicolumn{2}{c|}{{ImageNet-Sketch}} \\
     
     & {levels} & \texttt{acc@1} & \texttt{acc@5} 
     & \texttt{acc@1} & \texttt{acc@5} 
     & \texttt{acc@1} & \texttt{acc@5} \\
     
     \hline
     \hline
     
     {Dataset-specific models} & -
     & $65.50$ & $87.00$ 
     & $62.91$ & $83.66$ 
     & $67.36$ & $81.59$ \\ 
     {Multi-head model} & -
     
     & $65.12$ & $88.66$ 
     & $56.66$ & $81.37$ 
     & $38.07$ & $62.12$ \\

      %\textbf{Joint-head models} 
      Joint-head model & -
     & ${64.43}$ & ${88.48}$ 
     & ${55.52}$ & ${79.94}$ 
     & $63.40$ & $78.78$ \\
     \hline
      \multirow{2}{*}{Students (our)} & $\textbf{L}_1$ 
     & $68.81$          & $90.79$  
     & $\mathbf{66.78}$ & $\mathbf{87.10}$  
     & $69.31$          & $\mathbf{84.79}$  \\ 

    & $\textbf{L}_2$
     & $\mathbf{71.23}$ & $\mathbf{91.73}$  
     & $66.30$           & $86.84$  
     & $\mathbf{69.53}$ & $84.13$  \\ 
     
     \hline
     
     \multirow{2}{*}{Joint teacher (ours)} & $\textbf{L}_1$
     & $67.64$ & $\mathbf{89.81}$  %cifar100
     & $64.08$ & $\mathbf{85.44}$   %tiny-imagenet
     & $68.30$ & $\mathbf{82.90}$  \\ %ucf 

    & $\textbf{L}_2$
    & $\mathbf{70.16}$ & $88.99$  %cifar100
     & $\mathbf{65.02}$ & $84.08$  %tiny-imagenet
     & $\mathbf{68.35}$ & $82.69$  \\ %ucf 
     
     \hline
     
\end{tabular}
% }
\vspace{-0.2cm}
\caption{Results for the set $\mathcal{T}_2$, including the dataset-specific baselines, the students obtained by employing multi-level distillation using embeddings extracted at one ($\textbf{L}_1$) or two levels ($\textbf{L}_2$), and the corresponding joint teachers. The results of the best student and the best teacher are highlighted in bold.}
\label{tab:T2}
% \vspace{-0.2cm}
\end{table*}

\begin{figure*}[]
\centering

\includegraphics[width=0.75\textwidth]{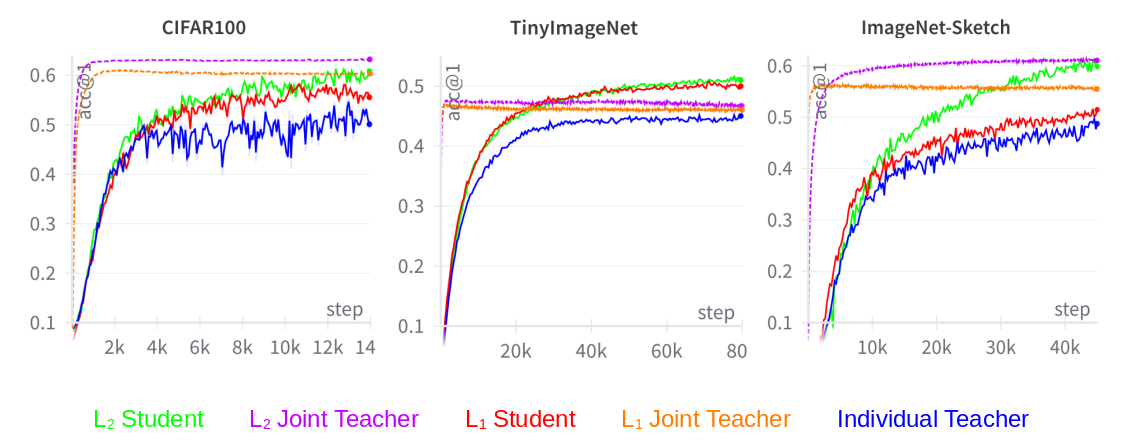}
\vspace{-0.2cm}
\captionof{figure}{Top-1 accuracy evolution during the training process for models in $\mathcal{T}_1$. Best viewed in color.}
\label{fig:results-T1}
\vspace{-0.2cm}
\end{figure*}

\subsection{Results}

We evaluate the performance of our approach in terms of two metrics: top-1 accuracy (\texttt{acc@1}) and top-5 accuracy (\texttt{acc@5}). We present quantitative results of our experiments in Table~\ref{tab:T1} with models from $\mathcal{T}_1$, and Table~\ref{tab:T2} with models from $\mathcal{T}_2$. Let $\textbf{L}_k = \{l_1, l_2, \ldots, l_k\}$ denote the set of $k$ representation levels used in our multi-level distillation scheme. Thus, $\textbf{L}_1$ and $\textbf{L}_2$ represent sets of one and two representation levels, respectively. 

\noindent{\bf {Students vs baselines}.} In terms of the top-1 accuracy, {\emph {all}} students surpass the performance of the dataset-specific models. The results are consistent over all three datasets and over all models in $\mathcal{T}_1$ and $\mathcal{T}_2$. The improvements range between 2\% and 12\%, depending on the number of representations levels. With respect to the dataset-specific models, our student models yield an average improvement of 3.4\% when the distillation uses a single representation level ($\textbf{L}_1$), and 6.1\% when the distillation is performed at two representation levels ($\textbf{L}_2$). We also consider two baselines jointly trained over all three datasets, as explained earlier. The multi-head multi-dataset model performs much worse than the dataset-specific models, while the joint-head multi-dataset model is comparable with the dataset-specific models. Both multi-dataset baselines are clearly outperformed by the students trained with our novel distillation method, with average improvements ranging between 5\% and 27\% in favor of our method.

\noindent{\bf {Importance of multi-level distillation}.} In five out of six cases, student models based on two-level distillation ($\textbf{L}_2$) outperform those based on one-level distillation ($\textbf{L}_1$), showing that employing knowledge distillation at multiple representations levels is beneficial. Notably, the students based on two-level distillation usually surpass the corresponding joint teacher models, with an average performance gain of 1\%.  

\begin{figure*}[!t]
\centering

\includegraphics[width=0.75\textwidth]{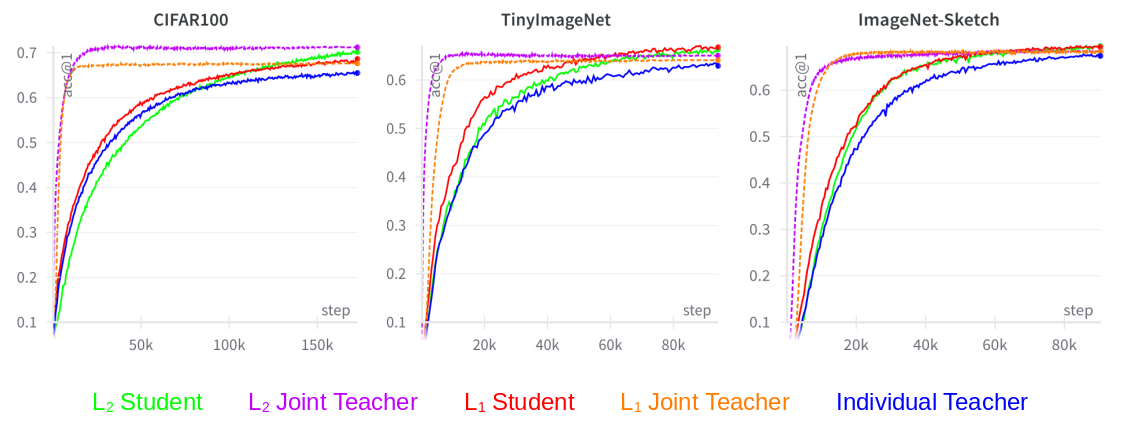}
\vspace{-0.2cm}
\caption{Top-1 accuracy during the training process for models in $\mathcal{T}_2$. Best viewed in color.}
\label{fig:results-T2}
\end{figure*}

\noindent{\bf {Performance evolution during training}.} Figures~\ref{fig:results-T1} and \ref{fig:results-T2} show the performance evolution of models from $\mathcal{T}_1$ and $\mathcal{T}_2$ during training, in terms of the top-1 accuracy. The plots show the accuracy on test data, after each training epoch. Both joint teachers converge very fast, while the corresponding students begin at the same pace with the individual teachers, but after a few epochs, they pick out the knowledge via the multi-level feature distillation method, and quickly catch up with the joint teachers. Interestingly, the student based on  $\textbf{L}_2$ is almost equal to its joint teacher and always outperforms the joint teacher based on $\textbf{L}_1$, showing, once again, the advantage of using multi-level representations in our framework. 

%In five out of six cases student models obtained when distillation uses two levels of representations ($L_2$ students) outperform students models learned using single level representations ($L_1$ students). This shows that employing knowledge in the form of features from multiple representations levels helps the distillation process. Moreover, the $L_2$ students perform on average a little better than the corresponding joint teacher models used for distillation ($L_2$ teachers), with about 1\% gain in performance.  

%and $L_2 = \{l_1,l_2\}$ the single and double -lvel representation depicted in the two tables.

\begin{table*}[t!]
% \vspace{-0.2cm}
\centering
%\resizebox{\columnwidth}{!}{%
\begin{tabular}{|c|c|cc|cc|cc|}
    \hline
      \multirow{2}{*}{{Model}} & {Representation} & \multicolumn{2}{c|}{{CIFAR-100}} & \multicolumn{2}{c|}{{TinyImageNet}} & 
     \multicolumn{2}{c|}{{ImageNet-Sketch}} \\
     
     & {levels} & \texttt{acc@1} & \texttt{acc@5} 
     & \texttt{acc@1} & \texttt{acc@5} 
     & \texttt{acc@1} & \texttt{acc@5} \\
     
     \hline
     \hline
     
     %{individual teacher (dataset-specific teacher)} 
     Dataset-specific models & -
     & $55.07$ & $81.62$ 
     & $45.09$ & $70.52$ 
     & $49.67$ & $71.39$ \\ 

      \hline
     
    \multirow{4}{*}{Joint teacher} & {$\textbf{L}_1$} 
     & $60.90$ & $84.56$  
     & $47.10$ & $72.43$  
     & $54.85$ & $75.20$  \\ 

     & {$\textbf{L}_2$}  
     & $\mathbf{62.28}$ & $\mathbf{85.75}$  
     & $\mathbf{47.52}$ & $\mathbf{73.24}$  
     & $\mathbf{61.81}$ & $\mathbf{78.94}$  \\ 

    & {$\textbf{L}_3$}  
     & $62.12$ & $85.71$  
     & $44.10$ & $70.40$  
     & $55.72$ & $76.28$  \\ 
     
     & {$\textbf{L}_4$}  
     & $48.65$ & $78.78$  
     & $29.93$ & $58.01$  
     & $30.69$ & $47.50$  \\ 
     
     \hline    
\end{tabular}
%}
\vspace{-0.2cm}
\caption{Results with multiple variants of joint teachers build with models from $\mathcal{T}_1$, obtained by varying the representation levels from $\textbf{L}_1$ to $\textbf{L}_4$.}
\label{tab:ablation}
\vspace{-0.2cm}
\end{table*}

\subsection{Ablation Study}

\begin{figure}[]
\centering
\includegraphics[width=1.\columnwidth]{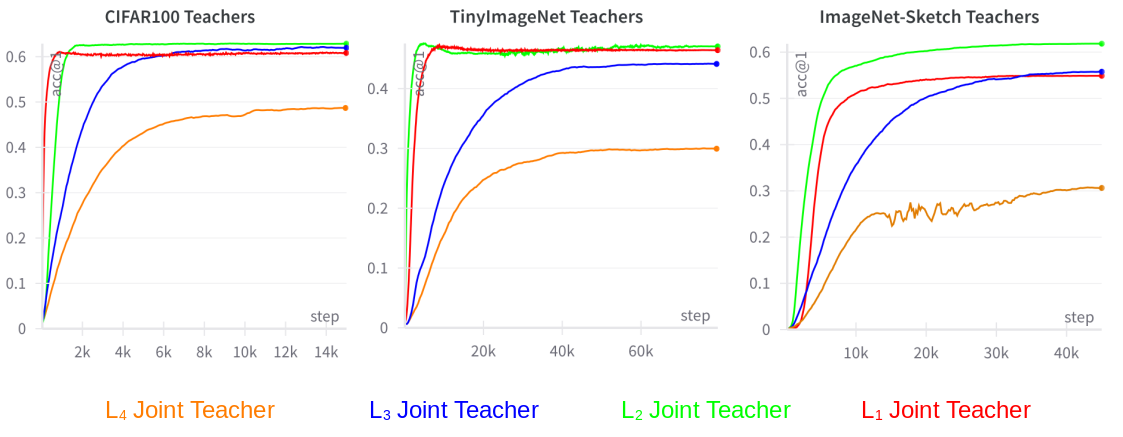}
\vspace{-0.2cm}
\caption{Performance evolution of joint teachers when using different sets of layers (from $\textbf{L}_1$ to $\textbf{L}_4$) to extract features. Best viewed in color.}
\label{fig:ablation}
\vspace{-0.35cm}
\end{figure}

We motivate the choice of the set of layer indexes $\textbf{L}_i$ at which the multi-level feature fusion and distillation is performed in our method through an ablation study, using models in the set $\mathcal{T}_1$. We evaluate the performance of the joint teachers considering representation levels in the following sets: $\textbf{L}_1$, $\textbf{L}_2$, $\textbf{L}_3$ and $\textbf{L}_4$. $\textbf{L}_1$ contains embedding vectors from the last layer. $\textbf{L}_2$ is obtained by adding feature maps of size $7 \times 7$ to $\textbf{L}_1$. Similarly, $\textbf{L}_3$ is obtained by adding feature maps of size $14 \times 14$ to $\textbf{L}_2$, and $\textbf{L}_4$ is obtained by adding feature maps of size $28 \times 28$ to $\textbf{L}_3$. As shown in Table~\ref{tab:ablation}, training joint teachers based on features extracted from layers closer to the input ($\textbf{L}_3$ or $\textbf{L}_4$) results in poor performance. Results are typically better when considering only a reduced number of layers, \ie~the last layer ($\textbf{L}_1$) or the last two layers ($\textbf{L}_2$) before the classification head. A possible explanation would be that layers farther from the output learn more generic (low-level) features, which do not bring new (dataset-specific) information into the joint teacher. Figure~\ref{fig:ablation} illustrates the evolution of the top-1 accuracy of the joint teachers learned in these four scenarios, confirming our observations.

\subsection{Supplementary}

In the supplementary material, we show that our method applies to a distinct collection of datasets, as well as a different task (action recognition). Moreover, we show that the feature space learned by our framework is more discriminative via a t-SNE visualization of the embeddings. We also discuss the limitations of our method.

\vspace{-0.1cm}
\section{Conclusion}
\label{sec:conclusions}
\vspace{-0.1cm}

In this paper, we presented a novel distillation method, coined Multi-Level Feature Distillation, that uses multiple teacher models combined into a joint teacher to distill the knowledge acquired at multiple representation levels from distinct datasets into student models. Our experimental evaluation for the task of image classification on three benchmark datasets demonstrated the benefits of our novel method in increasing the accuracy of several different state-of-the-art models, that are based on convolutional or transformer architectures. Additionally, we presented t-SNE visualizations that show the effectiveness of the knowledge distillation approach in learning robust latent representations, and explaining the superior results of our student models.

In future work, we aim to apply our method beyond image classification and explore new vision tasks, \eg~object detection and image segmentation, as well as new domains, \eg natural language processing.

%%%%%%%%% REFERENCES
{\small
\bibliographystyle{ieee_fullname}
\bibliography{egbib}
}

\clearpage
{\centering
        \Large
        \vspace{0.5em}\textbf{Supplementary Material} \\
        \vspace{1.0em}
}
        
\section{Overview}
\label{sec:addabl}

In order to offer a better understanding of the generality of our method, we include additional experiments. First of all, we aim to show that our method applies to a distinct collection of datasets. Second of all, we aim to quantify the impact of the number of datasets $m$ on the joint model, as well as on the distilled students. To this end, we follow the setup described in the main paper, but with the following differences: (i) we increase the number of datasets from three to four; (ii) we change the entire collection of datasets; (iii) we train the joint teacher by varying the number of datasets, from two to four. 

In another set of experiments, we compare with a single-dataset KD. This experiment aims to assess the benefits of distilling from multiple datasets.

We also aim to determine the applicability of our framework to distinct tasks. To this end, we conduct experiments on three action recognition datasets. In this setup, we start from pre-trained action recognition models, which allows us to demonstrate the effectiveness of our approach in conjunction with pre-training.

We further present ablation results where the joint teacher is either based on different backbones trained on the same dataset, or the same backbone trained on different datasets. Another dataset-related ablation is performed to demonstrate the generalization capacity of the MLFD students.

Next, we analyze the feature space generated by our approach. By visualizing the embeddings via t-SNE, we are able to determine that our approach leads to robust and disentangled representations. Finally, we discuss the time and space limitations of our framework.

\section{Additional Image Classification Results}

\subsection{Datasets}
\noindent
\textbf{Caltech-101.} 
The Caltech-101 dataset \cite{Fei-TPAMI-2006} consists of $7,315$ training images and $1,829$ test images. It was originally proposed to test the ability of models to learn from few examples. The images belong to 101 object categories.

\noindent
\textbf{Flowers-102.} The Flowers-102 dataset \cite{Nilsback-ICVGIP-2008} contains $7,169$ training images and $1,020$ test images. The dataset contains 102 classes of flowers that typically grow in the United Kingdom.

\noindent
\textbf{CUB-200-2011.} The Caltech-UCSD Birds 200 2011 (CUB-200) dataset \cite{Wah-CUB-2011} is formed of $9,430$ training images and $2,358$ test images. The images represent 200 distinct species of birds.

\noindent
\textbf{Oxford Pets.} 
The Oxford-IIIT Pets dataset \cite{Parkhi-CVPR-2012} consists of $5,906$ training images and $1,477$ test images. The dataset contains images for 37 breeds of cats and dogs.

\subsection{Models}

Since the datasets are distinct, we evaluate our method on a new set of individual teachers, denoted as $\mathcal{T}_3$, containing four models, namely a ResNet-18 \cite{he-2016-cvpr} trained on Caltech-101, an EfficientNet-B0 \cite{tan-2019-plmr} trained on Flowers-102, a SEResNeXt-26D \cite{hu-2018-cvpr} trained on CUB-200, and a ResNet-18 trained on Oxford Pets. $\mathcal{T}_3$ contains similar models to $\mathcal{T}_1$, but trained (from scratch) on distinct datasets.

\begin{table*}[]
\centering
% \resizebox{\columnwidth}{!}{%
\begin{tabular}{|c|c|cc|cc|cc|cc|}
    \hline
      \multirow{2}{*}{{Model}} & \multirow{2}{*}{\#Datasets} & \multicolumn{2}{c|}{{Caltech-101}} & \multicolumn{2}{c|}{{Flowers-102}} & 
     \multicolumn{2}{c|}{{CUB-200}} & \multicolumn{2}{c|}{{Oxford Pets}}  \\
     
     &  & \texttt{acc@1} & \texttt{acc@5} 
     & \texttt{acc@1} & \texttt{acc@5} 
     & \texttt{acc@1} & \texttt{acc@5} 
     & \texttt{acc@1} & \texttt{acc@5} \\
     
     \hline
     \hline
     
     {Dataset-specific models} & 1
     & $74.79$ & $86.39$ 
     & $78.04$ & $93.14$ 
     & $51.40$ & $77.06$ 
     & $63.91$ & $89.10$ \\ 
    \hline
    
      \multirow{3}{*}{Students (our)} & 2 
     & $79.66$          & $92.29$  
     & $80.78$          & $94.22$  
     & -          & - 
     & -          & - \\ 

    & 3
     & $\mathbf{80.32}$          & $91.31$ 
     & $79.41$          & $95.10$  
     & $58.27$          & $82.65$  
     & -              & - \\ 

     & 4
     & $80.04$            & $\mathbf{92.45}$  
     & $\mathbf{81.76}$   & $\mathbf{95.29}$  
     & $\mathbf{59.29}$   & $\mathbf{83.29}$  
     & $\mathbf{68.79}$   & $\mathbf{92.42}$  \\ 
     
     \hline
     
     \multirow{3}{*}{Joint teacher (ours)} & 2
     & $82.56$ & $93.00$  %cifar100
     & $80.39$ & $93.73$   %tiny-imagenet
     & - & -  
     & - & - \\ %ucf 
          
    & 3
    & $\mathbf{83.11}$ & $93.06$  %cifar100
    & $81.76$ & $\mathbf{94.61}$  %tiny-imagenet
    & $57.21$ & $\mathbf{81.30}$ 
    & -       & - \\ %ucf 

    & 4
    & $83.00$ & $\mathbf{93.71}$  %cifar100
    & $\mathbf{81.86}$ & $94.51$  %tiny-imagenet
    & $\mathbf{57.29}$ & $80.66$  
    & $\mathbf{66.08}$ & $\mathbf{89.91}$ \\ %ucf 
     \hline
     
\end{tabular}%
% }
\caption{Results for the set $\mathcal{T}_3$, including the dataset-specific baselines, the students obtained by employing multi-level distillation using embeddings extracted at two levels ($\textbf{L}_2$), and the corresponding joint teachers. The number of datasets used to train the joint teachers is gradually increased from two to four. The results of the best student and the best teacher on each dataset are highlighted in bold.}
\label{tab:T3}
\end{table*}

\begin{figure*}[]
  \centering
  \begin{tabular}{cc}
  \includegraphics[width=0.40\linewidth]{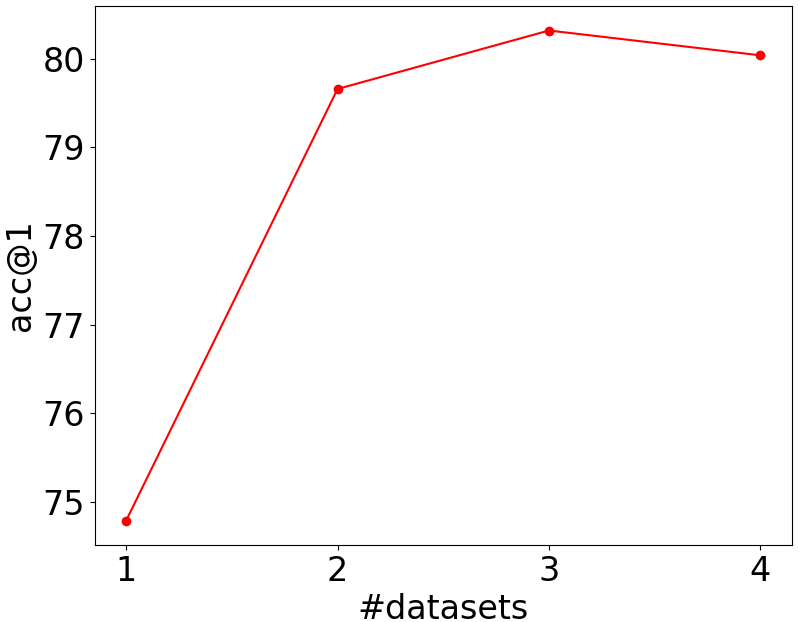} &
  \includegraphics[width=0.40\linewidth]{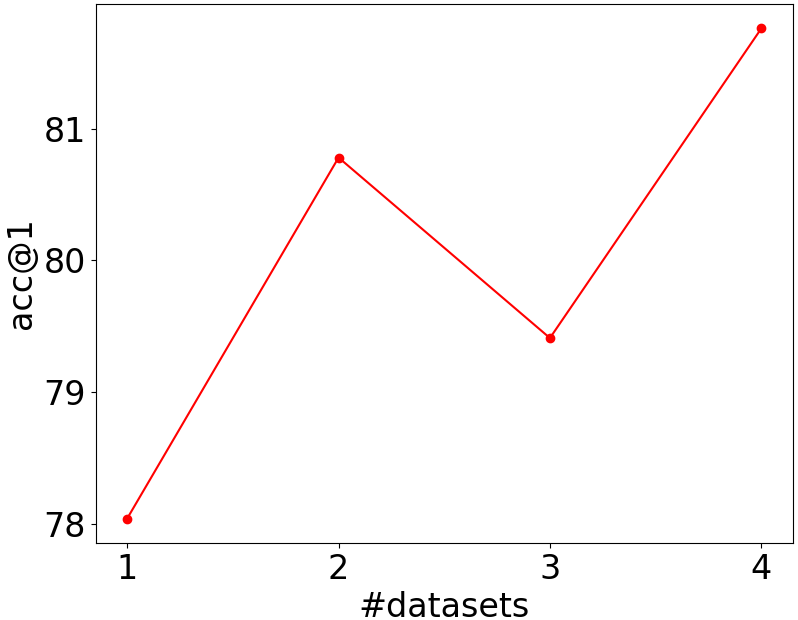} \\
  % (c) & (d) \\
  % (g) & (h) \\
  %(Left) & (Right)
  \end{tabular}
\vspace{-0.3cm}
\caption{Accuracy rates of the student models on Caltech-101 (left) and Flowers-102 (right) when the number of datasets is increased from one to four. Best viewed in color.}
\label{fig:ablation_datasets}
%\vspace{-0.3cm}
\end{figure*}

\subsection{Results}

Table~\ref{tab:T3} shows the results obtained when the joint teacher is trained on two, three and four datasets, respectively. The joint teacher is first optimized on Caltech-101 and Flowers-102. The next version adds the CUB-200-2011 dataset into the mix, and the last version is trained on all four datasets. Notably, the results indicate that two datasets are enough to reach substantial performance gains. We report additional gains when using more datasets, although the relative improvements tend to saturate with the number of datasets, as shown in Figure \ref{fig:ablation_datasets}. Regardless of the number of datasets, our multi-level distillation framework brings significant performance improvements on all four datasets. Even if the chosen datasets are typically small, the reported gains are still high, suggesting that our framework plays a very important role in increasing the performance and generalization capacity of neural models.

\begin{table*}[t!]
\setlength\tabcolsep{3.8pt}
    \centering
\begin{tabular}{|c|cc|cc|cc|}
    \hline
     \multirow{2}{*}{{Model}} & \multicolumn{2}{c|}{{CIFAR-100}} & \multicolumn{2}{c|}{{Tiny ImageNet}} & \multicolumn{2}{c|}{{ImageNet-Sketch}}  \\
     
     & \texttt{acc@1} & \texttt{acc@5} 
     & \texttt{acc@1} & \texttt{acc@5} 
     & \texttt{acc@1} & \texttt{acc@5} \\
     
     \hline
     \hline
     
     {Dataset-specific models (light=$\mathcal{T}_1$)}
     & $55.07$ & $81.62$ 
     & $45.09$ & $70.52$
     & $49.67$ & $71.39$ \\

     {Dataset-specific models (deep=$\mathcal{T}_4$)}
     & $64.38$ & $88.30$ 
     & $43.47$ & $69.48$
     & $57.75$ & $76.38$ \\ 
     \hline

     %\hline
      {Single-dataset KD (from $\mathcal{T}_1$ to $\mathcal{T}_1$)} 
     & $56.44$ & $82.63$ 
     & $48.15$ & $74.02$ 
     & $46.87$ & $70.40$ \\ 

        {Single-dataset KD (from $\mathcal{T}_4$ to $\mathcal{T}_1$)} 
     & $59.17$ & $83.68$ 
     & $47.16$ & $73.06$ 
     & $54.21$ & $75.45$ \\ 
     \hline
     
      {Multi-dataset KD (from $\mathcal{T}_1$ to $\mathcal{T}_1$)}
     %  & $\mathbf{58.65}$ & $\mathbf{83.97}$  
     % & $\mathbf{50.86 & 76.07  
     % & 51.64 & 72.81  \\ 
     
     & $\mathbf{62.25}$ & $\mathbf{84.64}$ %anet
     & $\mathbf{51.61}$ & $\mathbf{76.46}$ %hmdb
     & $\mathbf{61.31}$ & $\mathbf{78.36}$ \\ %ucf 
     
    % \textbf{State-of-the-art} \cite{Gowda2021AAAI, ryoo21arxiv} & $84.40$  & -- & -- &  \textbf{84.36} & -- & -- & \textbf{85.40} & -- & --
    % &  \textbf{98.64}  & -- & -- \\
     
     \hline
     
\end{tabular}
   \caption{Single-dataset KD (using individual dataset-specific models as teachers) versus our multi-dataset KD. Students are always light (and identical for both single-dataset and multi-dataset distillation). $\mathcal{T}_1$: ResNet-18, EfficientNet-B0, SEResNeXt-26D. $\mathcal{T}_4$: ResNet-50, EfficientNet-B1 and SEResNeXt-50D.}
    \label{tab_kd}
\end{table*}

\section{Comparison with Single-Dataset Distillation}

In Table \ref{tab_kd}, we present additional results with students based on standard (single-dataset) distillation. For the single-dataset distillation, we consider two distinct sets of teacher. On the one hand, we distill from the teachers included in $\mathcal{T}_1$, for a direct comparison with our multi-dataset approach. On the other hand, we use another set of more powerful teachers, namely $\mathcal{T}_4$, which is composed of the following models: ResNet-50 for CIFAR-100, EfficientNet-B1 for Tiny ImageNet, and SEResNeXt-50D for ImageNet-Sketch. The latter set of teachers is considered because it is common to use deeper teachers in teacher-student training setups. Nevertheless, the students trained with our multi-dataset distillation approach reach much better results than both single-dataset student versions. Our empirical results suggest that it is more effective to distill from lighter teachers trained on multiple datasets than distilling from a deeper teacher trained on a single dataset.

\section{Additional Action Recognition Results}

\begin{table*}[t!]
\setlength\tabcolsep{2.4pt}
    \centering
\begin{tabular}{|c|cc|cc|cc|}
    \hline
     \multirow{2}{*}{{Model}} & \multicolumn{2}{c|}{{ActivityNet}} & \multicolumn{2}{c|}{{HMDB-51}} & \multicolumn{2}{c|}{{UCF-101}}  \\
     
     & \texttt{acc@1} & \texttt{mAP} 
     & \texttt{acc@1} & \texttt{mAP} 
     & \texttt{acc@1} & \texttt{mAP} \\
     
     \hline
     \hline
     
     {Dataset-specific models}
     & $73.81$ & $42.83$ 
     & $73.60$ & $61.27$
     & $94.63$ & $86.32$ \\

     %\hline
      {Students} {($\textbf{L}_1$)} 
     & $\mathbf{81.56}$ & $\mathbf{82.89}$ 
     & $\mathbf{75.32}$ & $\mathbf{76.81}$ 
     & $\mathbf{95.97}$ & $\mathbf{97.96}$ \\ 
     
     \hline
     
      {Joint teacher} {($\textbf{L}_1$)} 
     & $88.50$ & $90.33$ %anet
     & $78.67$ & $79.13$ %hmdb
     & $98.05$ & $99.20$ \\ %ucf 

     \hline
     
\end{tabular}
   \caption{Action recognition results on ActivityNet, HMDB-51 and UCF-101 for a new set of models (ResNet-50 with Temporal Segment Network for ActivityNet; ResNet-50 with Temporal Shift Module for HMDB-51 and UCF-101), including the dataset-specific baselines, our students based on  $\textbf{L}_1$ embeddings, and the corresponding joint teacher. The dataset-specific models and individual teachers are pretrained on Kinetics-400. Our students outperform the dataset-specific models by large mAP gaps on all datasets.}
    \label{tab_ar}
\end{table*}

\begin{table*}[!t]
\centering
%\resizebox{\columnwidth}{!}{%
\begin{tabular}{|c|c|cc|cc|cc|}
    \hline
     \multirow{2}{*}{{Model}} & {Distillation} & 
     \multicolumn{2}{c|}{{TinyImageNet}} & 
     \multicolumn{2}{c|}{{ImageNet-Sketch}} \\
     
     & {level} 
     & \texttt{acc@1} & \texttt{acc@5} 
     & \texttt{acc@1} & \texttt{acc@5} \\
     
     \hline
     \hline
{Dataset-specific models} & - 
& $45.09$ & $70.52$ 
     & $49.67$ & $71.39$ \\ 
      %\hline
     \multirow{1}{*}{Students (same dataset)} 

      %\textbf{$L_2$ Joint Teacher}
     & $\textbf{L}_2$ 
     & $50.07$ & $75.45$  %tiny-imagenet
     & $54.16$ & $75.55$  \\ %ucf 

     %\hline
     
    \multirow{1}{*}{Students (different datasets)}

     %\textbf{$L_2$ Students} 
     & $\textbf{L}_2$ 
     & $\mathbf{51.61}$ & $\mathbf{76.46}$  
     & $\mathbf{61.31}$ & $\mathbf{78.36}$ \\
     \hline
     
\end{tabular}
%}
\caption{Comparison between students distilled from joint teachers of identical capacity, but using same or distinct datasets. The same-dataset students are distilled from a joint teacher that combines different backbones which are all trained on the target dataset. The students trained on different datasets are based on our unmodified MLFD framework.}
\label{tab:ensemble}
\end{table*}

\subsection{Datasets}
\noindent
\textbf{ActivityNet.} The ActivityNet dataset \cite{ActivityNet} comprises $19,994$ videos labeled with $200$ activity classes. Following standard evaluation practices, we report results on the official validation set, since there are no publicly-available labels for the test set.

\noindent
\textbf{HMDB-51.} The HMDB-51 dataset \cite{HMDB51} consists of $7,000$ clips distributed in $51$ action classes. The official evaluation procedure uses three different data splits. We consider the first split in our experiments.

\noindent
\textbf{UCF-101} The UCF-101 dataset \cite{UCF101} contains $13,320$ YouTube videos from $101$ action classes. As for HMDB-51, there are three data splits and we select the first one for our evaluation.

\subsection{Models}

The action recognition models are taken from the MMAction2\footnote{\url{https://github.com/open-mmlab/mmaction2}} toolbox, which provides various models pre-trained on different datasets. For ActivityNet, the individual teacher is a pre-trained Temporal Segment Network \cite{TSN} architecture, which is based on a ResNet-50 backbone with $8$ segments. For HMDB-51 and UCF-101, the teachers are based on a pre-trained ResNet-50 with Temporal Shift Module \cite{TSM}. All selected teachers are first pre-trained on the Kinetics-400 dataset \cite{K400}. Then, each teacher is fine-tuned on its own target dataset. From this point on, we employ our multi-dataset distillation approach.

\subsection{Results}

%experimente pe 10-20 de clase din CIFAR-100, vizualizare in 2D cu t-SNE din embedding space orginal vs cel dupa distilare

\begin{table*}[!h]
\centering
%\resizebox{\columnwidth}{!}{%
\begin{tabular}{|c|c|cc|cc|cc|}
    \hline
     \multirow{2}{*}{{Model}} & {Distillation} & 
     \multicolumn{2}{c|}{{Caltech-101}} & 
     \multicolumn{2}{c|}{{Oxford Pets}} \\
     
     & {level} 
     & \texttt{acc@1} & \texttt{acc@5} 
     & \texttt{acc@1} & \texttt{acc@5} \\
     
     \hline
     \hline

      % \hline
     { Dataset-specific models} 
     & - 
     & $74.79$ & $86.39$ 
     & $63.91$ & $89.10$ \\ 
      %\textbf{$L_2$ Joint Teacher}
      { Students (same architecture) }
     & $\textbf{L}_2$ 
     & $\mathbf{80.75}$ & $\mathbf{92.56}$  %tiny-imagenet
     & $\mathbf{68.99}$ & $\mathbf{93.09}$  \\ %ucf 

     \hline
     
    \multirow{1}{*}{Joint teacher (same architecture)}

     %\textbf{$L_2$ Students} 
     & $\textbf{L}_2$ 
     & $82.61$ & $93.27$  
     & $64.92$ & $89.10$ \\
     \hline
     
\end{tabular}
%}
\caption{Results with a joint teacher based on the same architecture (ResNet-18) trained on different datasets (Caltech-101 and Oxford Pets). The corresponding students are also based on ResNet-18.}
\label{tab:same-arch}
\end{table*}

We present action recognition results with $\textbf{L}_1$ students in Table \ref{tab_ar}. Although we start from pre-trained individual teachers, the joint teacher leads to significant performance gains. Distilling knowledge from the joint teacher into the student models is also beneficial. In the end, we obtain student models that are identical in terms of architecture to the dataset-specific models, but the action recognition performance of our students is significantly higher, especially in terms of mAP.

\section{Additional Ablations}

\begin{table}[!t]
\centering 
\begin{tabular}{|c|cc|}
    \hline
      \multirow{2}{*}{{Model}}& \multicolumn{2}{c|}{{Oxford Pets}}  \\
      
     & \texttt{acc@1} & \texttt{acc@5} \\
     
     \hline
     \hline
     
     {Dataset-specific model}
     & $63.91$ & $89.10$ \\ 
     \hline
    
    \multirow{1}{*}{Student \textbf{w/o} Oxford Pets}
     & $\mathbf{66.14}$   & $\mathbf{91.94}$  \\  
    %\hline
     
    \hline
     
\end{tabular}%
%\vspace{-0.5em}
\caption{Results on Oxford Pets with the dataset-specific model versus an $\textbf{L}_2$ student distilled from a joint teacher which is trained on three datasets: Caltech-101, Flowers-102, and CUB-200.}
\label{tab:3dist}
\end{table}

\begin{figure*}[!t]
  \centering
  \begin{tabular}{cc}
  \includegraphics[width=0.450\linewidth]{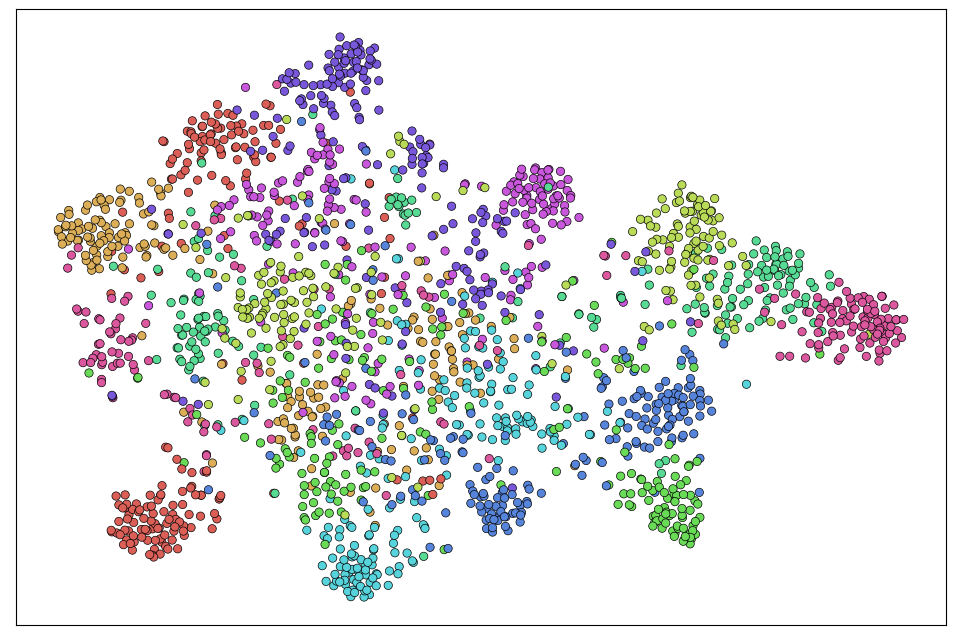} &
  \includegraphics[width=0.450\linewidth]{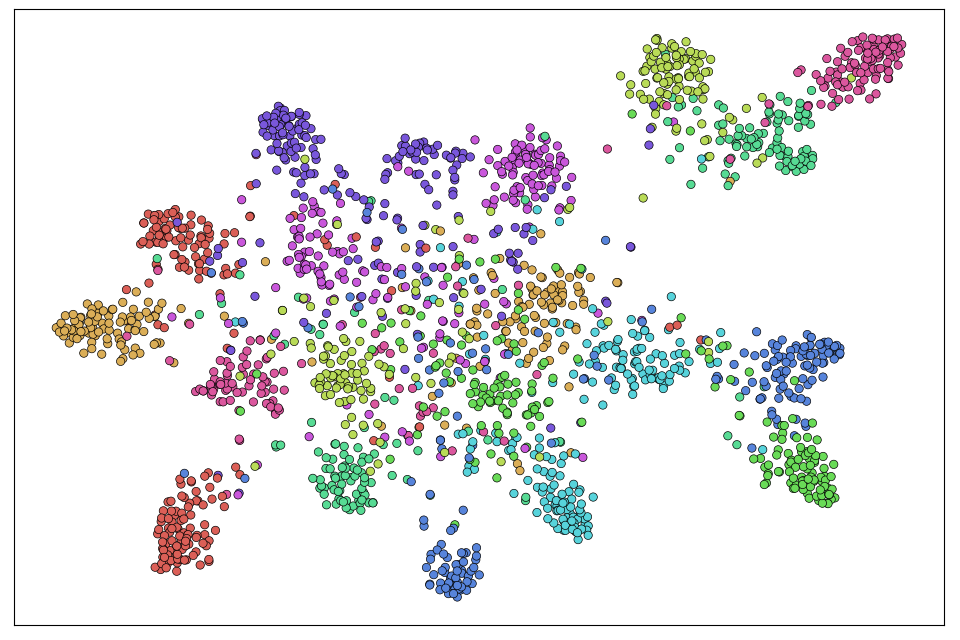} \\
  % (c) & (d) \\
  \includegraphics[width=0.450\linewidth]{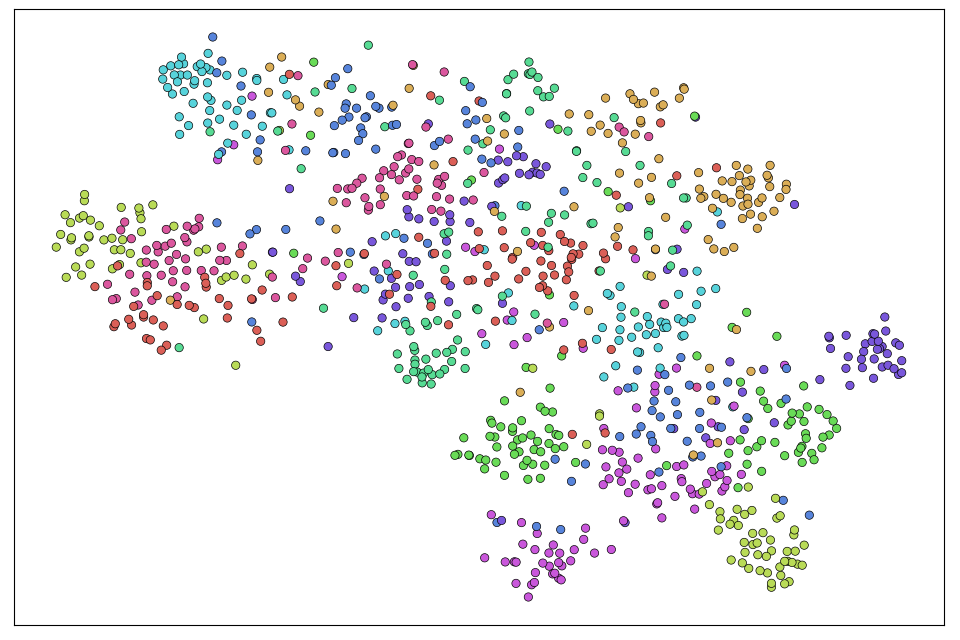} &
  \includegraphics[width=0.450\linewidth]{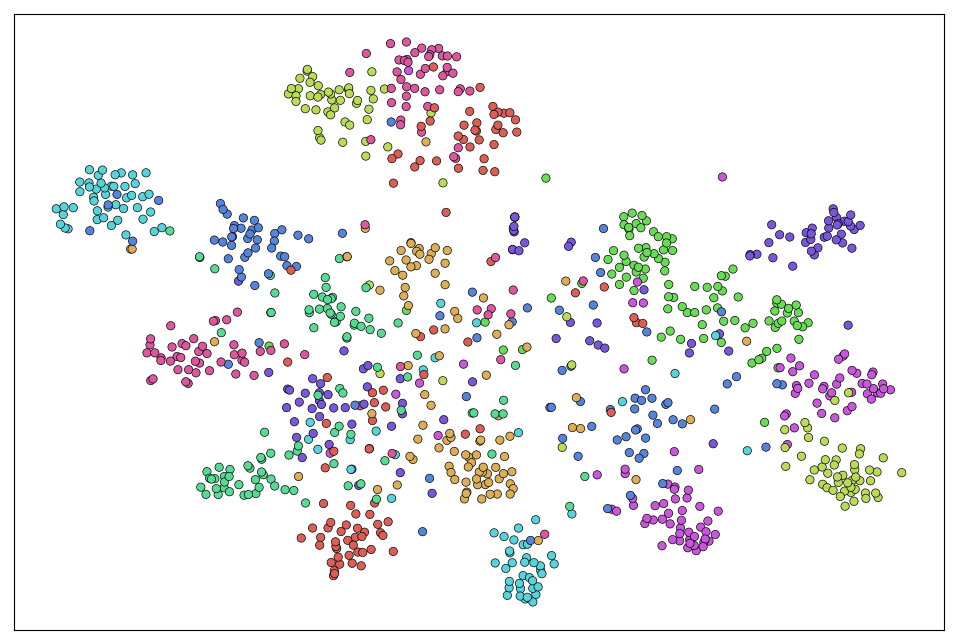} \\
  % (e) & (f) \\
  \includegraphics[width=0.450\linewidth]{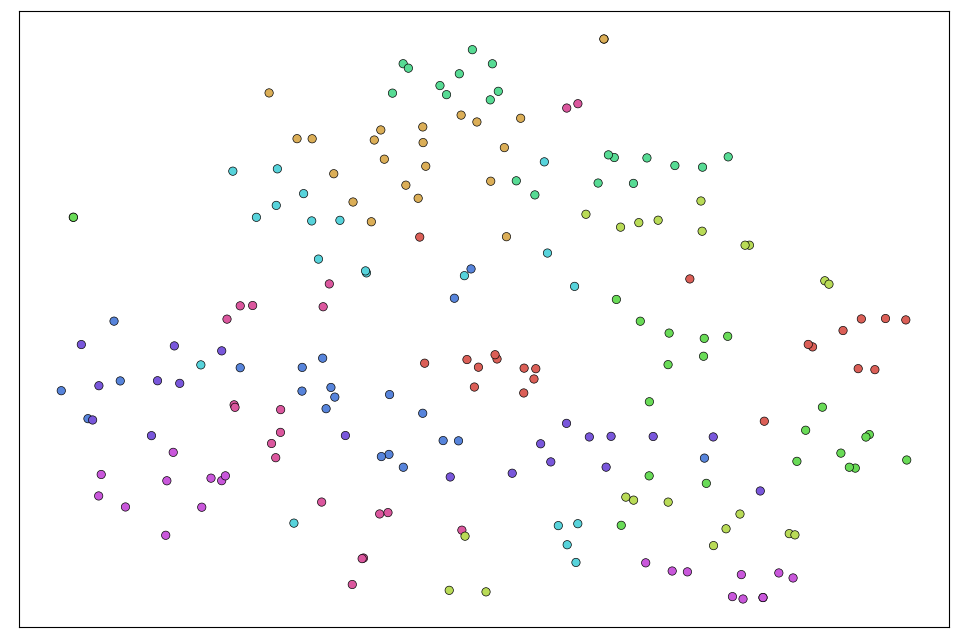} &
  \includegraphics[width=0.450\linewidth]{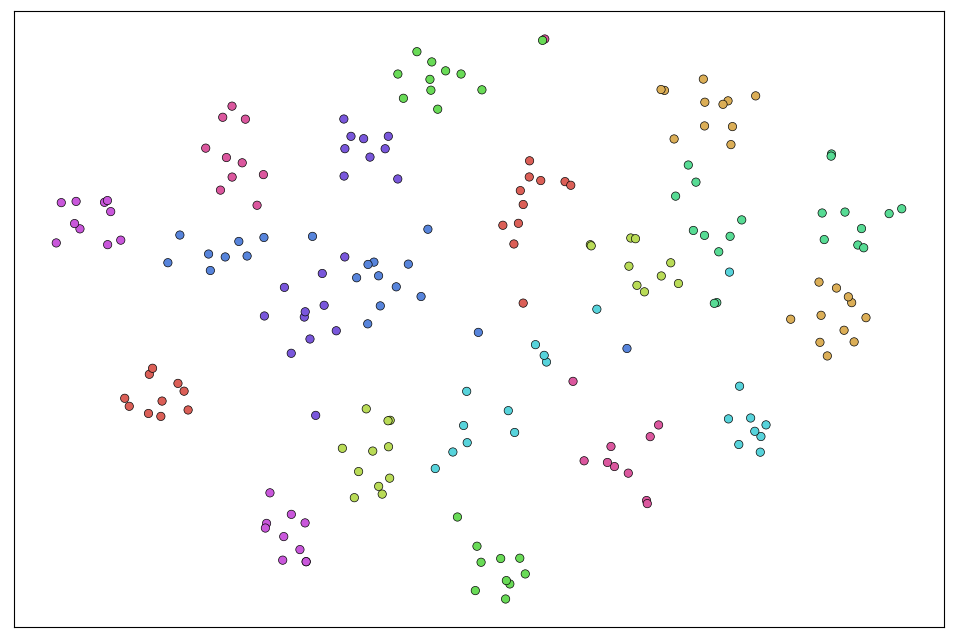} \\
  % (g) & (h) \\
  %(Left) & (Right)
  \end{tabular}
\vspace{-0.2cm}
\caption{Visualizations based on t-SNE projections of image embeddings learned by the dataset-specific models (left) and those learned by our student models (right) for the three datasets: CIFAR-100 (first row), TinyImageNet (second row), ImageNet-Sketch (third row). Best viewed in color.}
\label{fig:tsne}
\vspace{-0.3cm}
\end{figure*}

\subsection{Distillation from Same-Dataset Joint Teacher}

To demonstrate the utility of training the joint teacher on a diversity of datasets, we perform an ablation study where the joint teacher is based on the same variety of architectures, but all teachers are trained on the same dataset. In Table \ref{tab:ensemble}, we compare the students based on $\textbf{L}_2$ distillation for $\mathcal{T}_1$ teachers on TinyImageNet and ImageNet-Sketch. Although both kinds of students surpass the dataset-specific models, our multi-dataset students clearly benefit from the more diverse datasets used to train the joint teacher. The results are consistent on both TinyImageNet and ImageNet-Sketch.

\subsection{Distillation from Same-Architecture Joint Teacher}

To show that our multi-dataset distillation works even if the joint teacher uses the same architecture across different datasets, we perform an experiment where the joint teacher comprises a ResNet-18 trained on Caltech-101, and a ResNet-18 trained on Oxford Pets. The corresponding students are also based on ResNet-18. We report the results of the dataset-specific models, the joint teacher and the $\textbf{L}_2$ students in Table \ref{tab:same-arch}. Both teacher and student models outperform the dataset-specific models, confirming that our multi-dataset distillation performs well, even when the same architecture is employed across all datasets.

\subsection{Cross-Dataset Generalization}

To showcase the generalization capacity of our framework, we present cross-dataset results by training a joint teacher on Caltech-101, Flowers-102, CUB-200 and distilling the knowledge into an $\textbf{L}_2$ student for Oxford Pets. In Table \ref{tab:3dist}, we compare this student with the dataset-specific model on Oxford Pets. Our cross-dataset student surpasses the dataset-specific model, thus showing a higher generalization capacity.

\section{Feature Analysis}

To gain a deeper understanding of our method, we compare the discriminative power of the embeddings learned by the dataset-specific models and those learned by our student models. We achieve this using the t-SNE~\cite{maaten-2008-jmlr} visualization tool and plot the 2D projections of test data points (images) from different classes, as obtained after applying the corresponding embedding (model). As there are at least $100$ classes in each of the three considered datasets in the main paper, we plot the projections of test data points for only a fraction of the total number of classes, to improve clarity. Figure~\ref{fig:tsne} shows  the 2D projections obtained using the t-SNE tool, for each of the three datasets. The visualization reveals that the embeddings learned by our student models cluster data points from the same class much better than the dataset-specific models, thus demonstrating a higher discriminative power. 

\vspace{-0.1cm}
\subsection{Limitations}
\label{sec:limitations}
% In this section we discuss some of the limitations of our proposed method.
\vspace{-0.1cm}

\noindent\textbf{Time complexity.} A possible limitation of our method is the wall-clock training time. 
The number of models that need to be trained is proportional with the number of datasets $m$. The $m$ individual teachers specific to each dataset can be trained in parallel, whereas the joint teacher and the student models need to be trained sequentially. In general practice, the training time can also be reduced by using pre-trained networks as individual teachers. 
%For this method, we have as limitations the increasing of training time.
%Assuming the training time for a regular model to be $\mathbf{N}$ seconds, then the training time becomes at most $3 \times \mathbf{N}$ seconds, due to time of training the joint teacher and then retraining the initial model. Our experimental evaluation shows in Figure~\ref{fig:results-T1} and Figure~\ref{fig:results-T2} that the joint teachers tend to converge significantly faster than the initial models, achieving better performances after fewer training iterations, in both L1 and L2 joint teachers setups. 
In our experiments, the joint teachers obtain stable performance after roughly 1/5 of the training time of the individual teachers, since they can harness the information learned by individual teachers.  
% 1/7, 1/8, and 1/5  of the training time of the initial model. 
Based on these insights, the total training time required to obtain a student ranges between $1.2$ (if all individual teachers are pre-trained) and $m+1.2$ (if all individual teachers need to be trained), where $m$ is the number of data sets. Notice that, during inference, the wall clock-time remains the same, \ie there is no difference between the dataset-specific models and our students.

\noindent
\textbf{Space complexity.}
Regarding space complexity, the only limitation is the storage required for caching latent representations of the individual teachers, especially when $m$ is larger. In practice, we can limit the number of individual teacher models and datasets to a manageable size, \eg 2-4, to avoid using too much storage space. Our results show that $m=3$ is enough to bring significant performance gains, up to $12\%$.

\end{document}